\documentclass[10pt,twocolumn,letterpaper]{article}

\usepackage{times}
\usepackage{epsfig}
\usepackage{graphicx}
\usepackage{amsmath}
\usepackage{amssymb}
\usepackage{url}

\begin{document}

\title{Segmenting Dermoscopic Images}

\author{Mario~Rosario~Guarracino, Lucia~Maddalena\\
National Research Council, Institute for High-Performance Computing and Networking\\
Via P. Castellino 111, 80131 Naples, Italy\\
{\tt\small \{mario.guarracino,lucia.maddalena\}@cnr.it} }

\date{}

\maketitle

\begin{abstract}
We propose an automatic algorithm, named SDI, for the segmentation
of skin lesions in dermoscopic images, articulated into three main
steps: selection of the image ROI, selection of the segmentation
band, and segmentation.
We present extensive experimental results achieved by the SDI
algorithm on the lesion segmentation dataset made available for
the ISIC 2017 challenge on Skin Lesion Analysis Towards Melanoma
Detection, highlighting its advantages and disadvantages.
\end{abstract}

\section{Introduction}

Several diagnostic protocols are usually adopted by dermatologists
for analyzing and classifying skin lesions, such as the so-called
\emph{ABCD-rule} of dermoscopy \cite{Stol94}.
Due to the subjective nature of examination, the accuracy of
diagnosis is highly dependent upon human vision and
dermatologist's expertise.
%
%
Computerized dermoscopic image analysis systems, based on a
consistent extraction and analysis of image features, do not have
the limitation of this subjectivity. These systems involve the use
of a computer as a second independent and objective diagnostic
method, which can potentially be used for the pre-screening of
patients performed by non-experienced operators. Although
computerized analysis techniques cannot provide a definitive
diagnosis, they can improve biopsy decision-making, which some
observers feel is the most important use for dermoscopy
\cite{Burroni04}.
Recently, numerous researches on this topic propose systems for
the automated detection of malignant melanoma in skin lesions
(e.g., \cite{Celebi07,Celebi09,Maglogiannis09,Cozza2011,Celebi2015,Celebi2015b}).
In our previous study on dermoscopic images \cite{Cozza2011}, the
segmentation of the skin area and the lesion area was achieved by
a semi-automatic process based on Otsu algorithm \cite{Otsu1979},
supervised by a human operator. Here, we propose a full automatic
segmentation method consisting of three main steps: selection of
the image ROI, selection of the segmentation band, and
segmentation.

The paper is organized as follows.
In Section \ref{ProposedApproach} we describe the proposed
algorithm, providing details of its main steps.
In Section \ref{ExpRes} we provide a thorough analysis of
experimental results on the ISIC 217 dataset \cite{ISICdataset}.
Conclusions are drawn in Section \ref{Conclusioni}.

\section{SDI Algorithm} \label{ProposedApproach}

\begin{figure}[!ht]
\begin{center}
\includegraphics[width=0.99\linewidth]{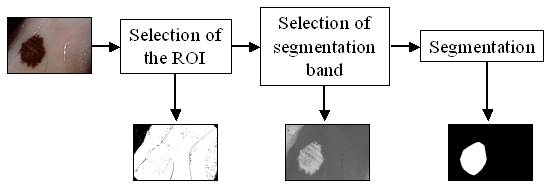}
\end{center}
\caption{Block diagram of the SDI algorithm.} \label{Fig:Overall}
\end{figure}

The block diagram of the segmentation algorithm proposed for
dermoscopic images, named SDI algorithm, is shown in Fig.
\ref{Fig:Overall}. The three main steps are described in the
following.

\subsection{Selection of the Image ROI}

In order to achieve an easier and more accurate segmentation of
the skin lesion, it is advisable to select the region of interest
(ROI), i.e., the subset if image pixels that belong to either the
lesion or the skin. This region excludes image pixels belonging to
(usually dark) areas of the image border and/or corners, as well
as those belonging to hair, that will not be taken into account in
the subsequent steps of the SDI algorithm.

In the proposed approach, the Value band of the image in the HSV
color space is chosen in order to select dark image pixels; these
are excluded from the ROI if they cover most of the border or the
angle regions of the image.

Concerning hair, many highly accurate methods have been proposed
in the literature \cite{Celebi2015}. Here, we adopted a bottom-hat
filter in the Red band of the RGB image.

\begin{figure}[!ht]
\begin{center}
\begin{tabular}{cc}
\includegraphics[width=0.28\linewidth]{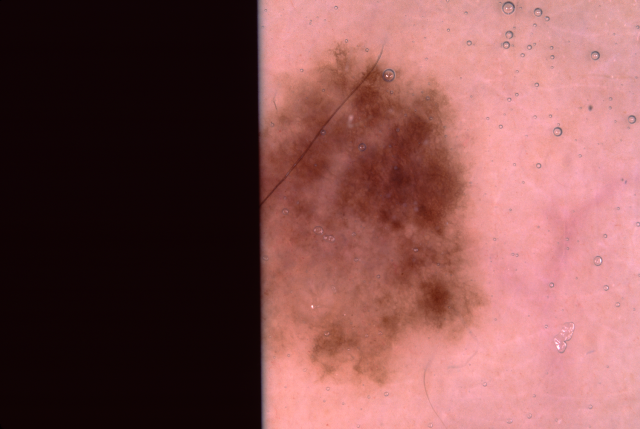} &
\includegraphics[width=0.28\linewidth]{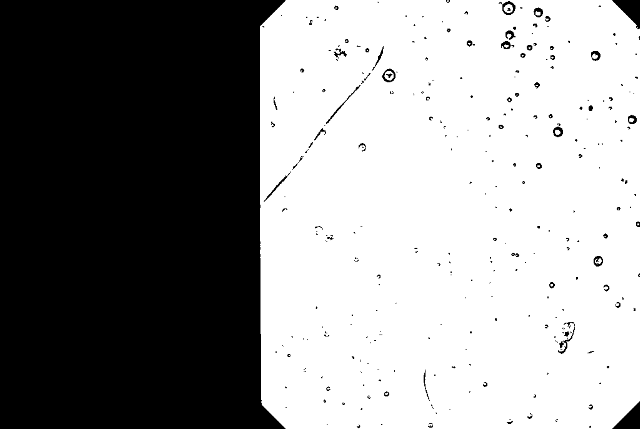}\\
(a) & (b)\\
\end{tabular}
\end{center}
\caption{Selection of the ROI (b) for image 15544 (a).}
\label{Fig:ROIselection}
\end{figure}

An example of the ROI selection process is reported in Fig.
\ref{Fig:ROIselection} for the ISIC 2017 test image no. 15544.
Here, we observe that the wide dark border on the left of the
image, as well as the dark hair over the lesion, have properly
been excluded from the ROI mask.

\subsection{Selection of the Segmentation Band}

Lesion segmentation can be made easier if the proper color band of the dermoscopic image is chosen. After thorough experimentation, we selected two color bands that allow proper segmentation: the Red band (Rnorm) in the normalized RGB color space and the Value band (V) of the image in the HSV color space.

\begin{figure}[!htb]
\setlength{\tabcolsep}{2pt} \centering
\begin{tabular}{ccccc}
\includegraphics[width=0.18\linewidth]{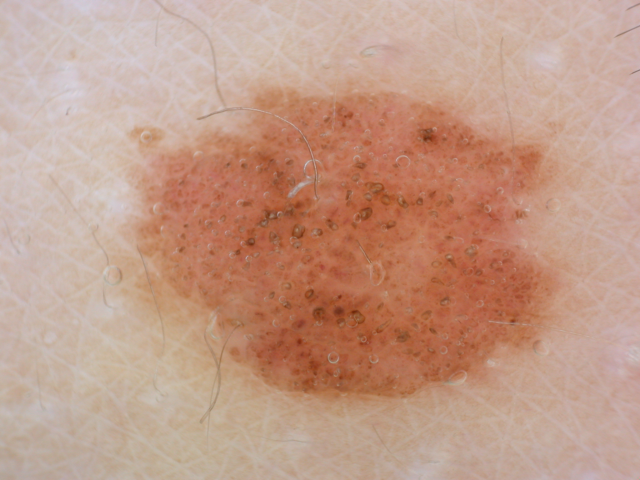}         &
\includegraphics[width=0.18\linewidth]{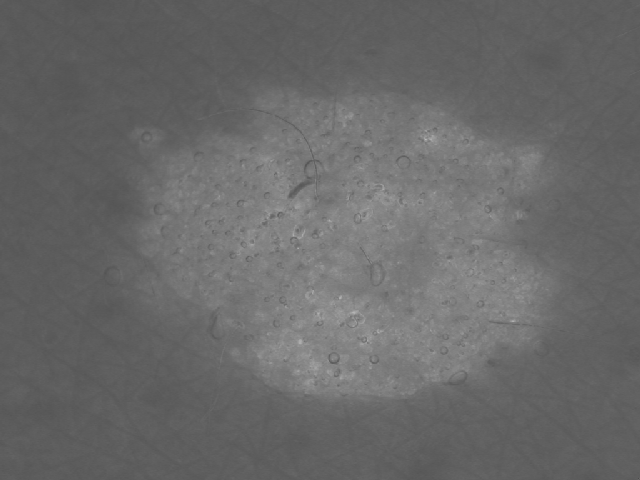}            &
\includegraphics[width=0.18\linewidth]{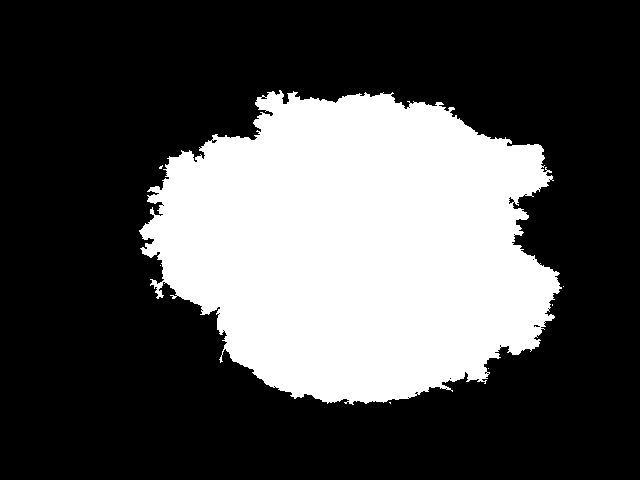}     &
\includegraphics[width=0.18\linewidth]{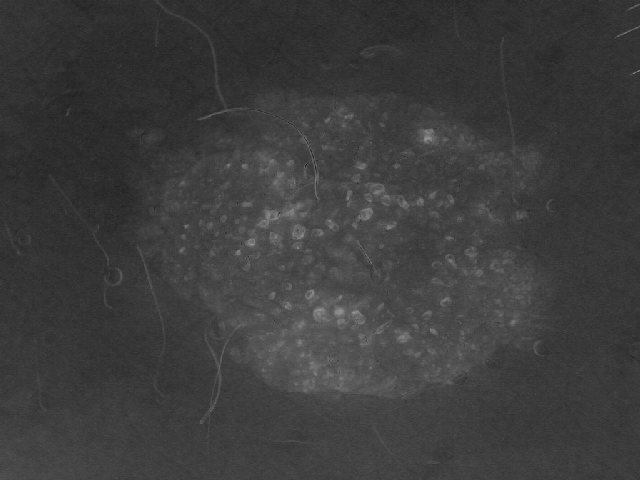}            &
\includegraphics[width=0.18\linewidth]{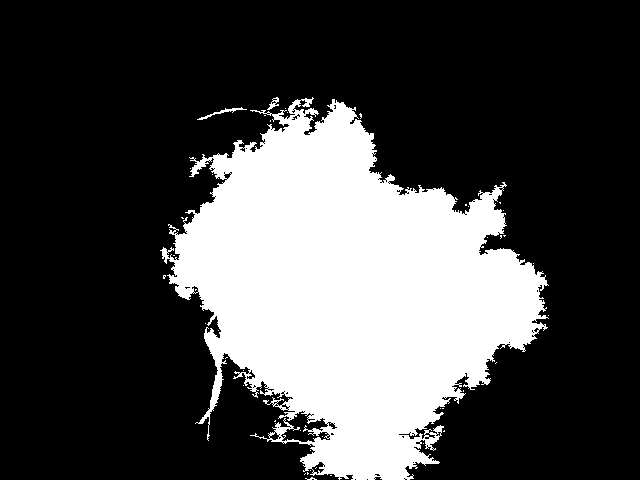}\\
\includegraphics[width=0.18\linewidth]{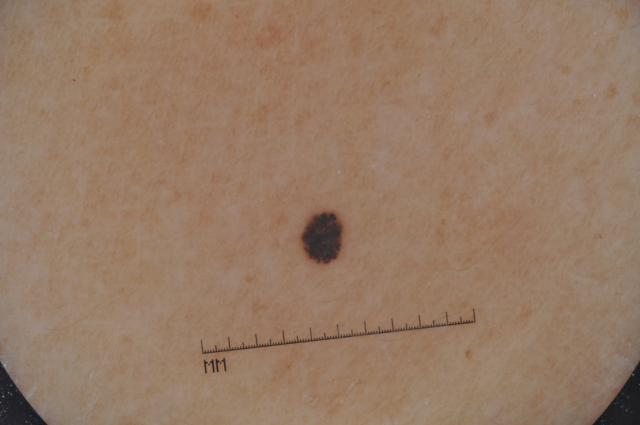}       &
\includegraphics[width=0.18\linewidth]{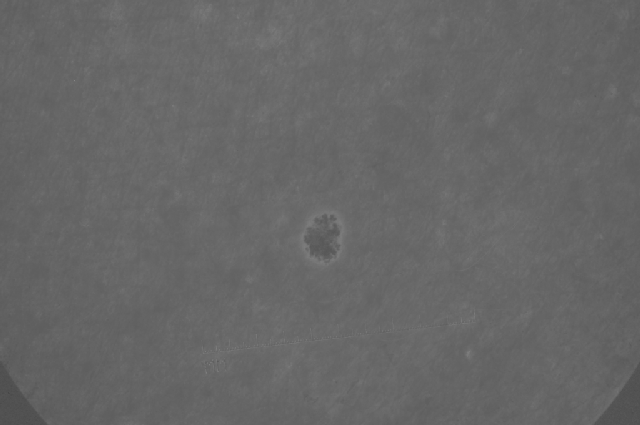}              &
\includegraphics[width=0.18\linewidth]{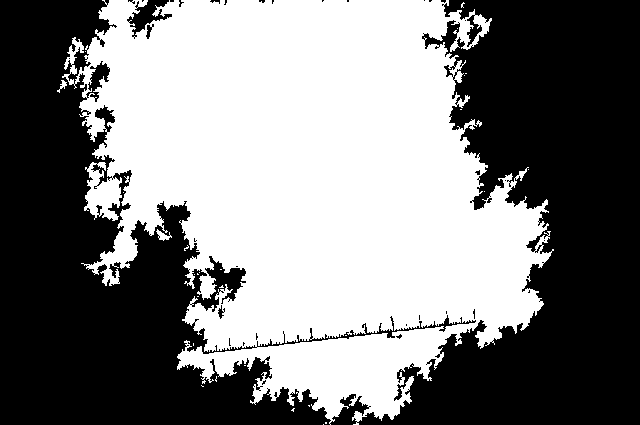}   &
\includegraphics[width=0.18\linewidth]{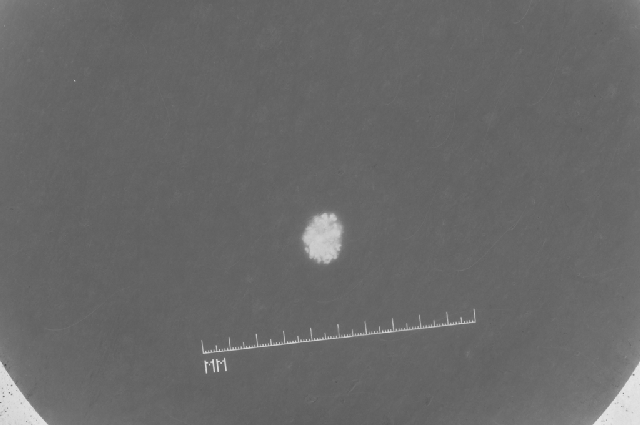}          &
\includegraphics[width=0.18\linewidth]{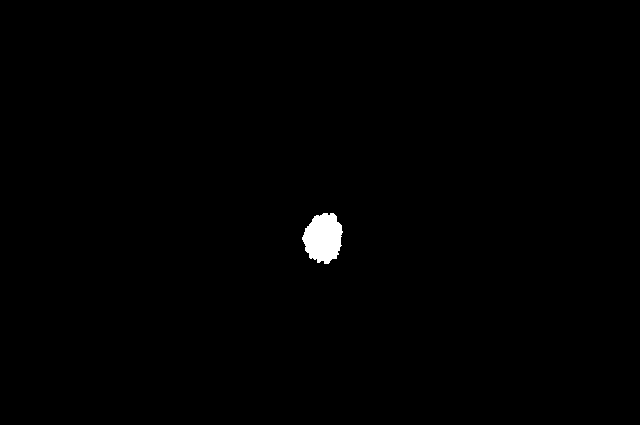}\\
(a) & (b) & (c) & (d) & (e)\\
\end{tabular}

\caption{Selection of the segmentation band for images 122 (first
row) and 12481 (second row): (a) color image; (b) Red band in the
normalized RGB color space; (c) binarization of (b); (d) inverted
Value band of the image in the HSV color space; (e) binarization
of (d).} \label{Fig:Channelselection122}
\end{figure}

Indeed, Rnorm is often a good choice for segmentation of
dermoscopic images, as the normalized RGB space eliminates the
effect of varying intensities due to uneven illumination and it is
free from shadow and shading effects. For example, for ISIC
training image 122 (first row of Fig.
\ref{Fig:Channelselection122}-(a)) the Rnorm band (Fig.
\ref{Fig:Channelselection122}-(b)) is not affected by the uneven
illumination of the image (brighter in the upper part), and its
binarization (Fig. \ref{Fig:Channelselection122}-(c)) provides a
quite faithful lesion segmentation. Instead, the V band (Fig.
\ref{Fig:Channelselection122}-(d)) is affected by the uneven
illumination and its binarization (Fig.
\ref{Fig:Channelselection122}-(e)) includes into the segmentation
mask also scarcely illuminated skin areas in the bottom of the
lesion.

However, there are cases where the V band is a better choice, as shown for ISIC training image 12481 (second row of Fig. \ref{Fig:Channelselection122}). Here, the Rnorm band almost annihilates the discrimination of the lesion by the surrounding skin, leading to a wrong segmentation, while thresholding in V band provides an almost perfect segmentation.

A comparison of the segmentations provided by the two color bands allows us to automatically select the most appropriate for the final segmentation.

\subsection{Segmentation}

Once the proper color band of the image has been selected,
segmentation is achieved by the Otsu algorithm, that computes the
optimal threshold separating the two classes of pixels (skin and
lesion) so that their intra-class variance is minimal
\cite{Otsu1979}.

The lesion area is then selected in the obtained binary mask as
the connected component having maximum area. This choice is based
on the assumption that dermoscopic images tend to mainly frame the
lesion to be analyzed, that thus appear in the image as the
predominant objects over the patient skin.

The convex hull of the segmented lesion is then adopted as final
segmentation result. Indeed, although the segmented lesion better
highlights the lesion contours, generally its convex hull better
conforms to the ground truth provided by the dermatologist.

\begin{figure}[!htb]
\setlength{\tabcolsep}{2pt} \centering
\begin{tabular}{cccc}
\includegraphics[width=0.18\linewidth]{122.png}        &
\includegraphics[width=0.18\linewidth]{ISIC_0000122BW1.png} &
\includegraphics[width=0.18\linewidth]{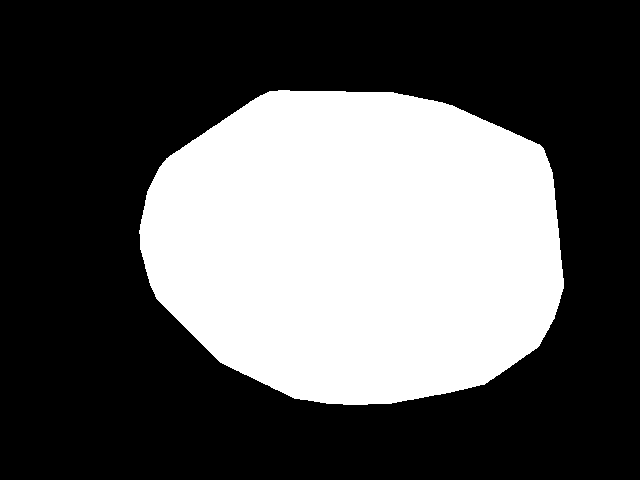} &
\includegraphics[width=0.18\linewidth]{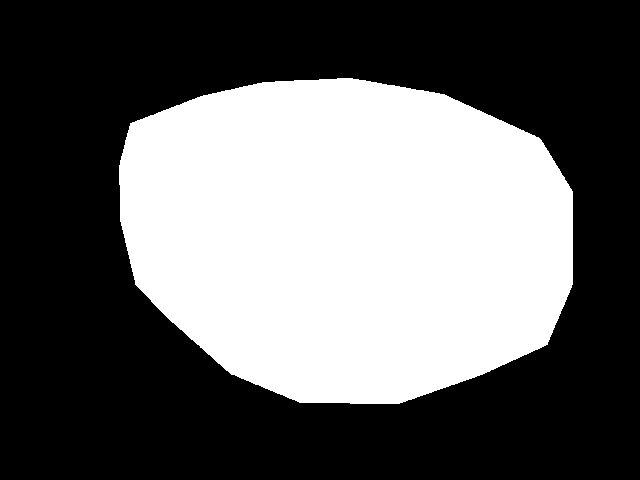} \\
(a) & (b) & (c) & (d)\\
\end{tabular}

\caption{Segmentation: (a) image 122; (b) initial SDI
segmentation; (c) final SDI segmentation; (d) ground truth.}
\label{Fig:SegmentationCH}
\end{figure}

An example of the segmentation process is reported in Fig.
\ref{Fig:SegmentationCH} for the ISIC 2017 train image no. 122.
Here, we observe that the initial SDI segmentation (the connected
component having maximum area, reported in Fig.
\ref{Fig:SegmentationCH}-(b)) provides a quite faithful
segmentation of the image lesion (Fig.
\ref{Fig:SegmentationCH}-(a)). The final SDI segmentation (the
convex hull, reported in Fig. \ref{Fig:SegmentationCH}-(c)) gives
a lesion segmentation that is much rougher, but more similar to
the ground truth mask (Fig. \ref{Fig:SegmentationCH}-(d)).

\section{Analysis of Experimental Results} \label{ExpRes}

Here we analyze in detail some of the results achieved by the SDI
algorithm on the test segmentation set of the ISIC 2017 challenge,
highlighting pro's and con's.

\begin{figure}[!htb]
\setlength{\tabcolsep}{2pt} \centering
\begin{tabular}{ccccc}
\includegraphics[width=0.18\linewidth]{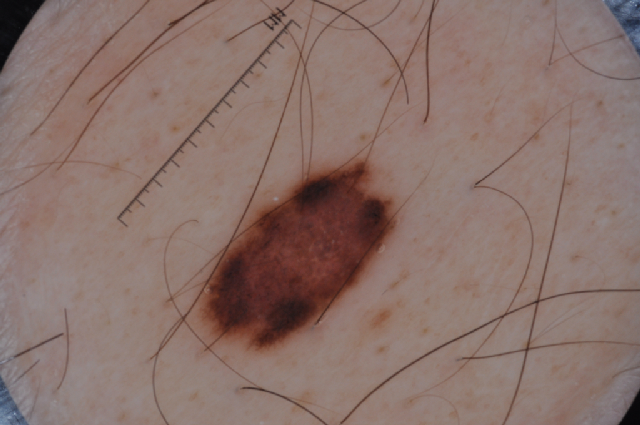} &
\includegraphics[width=0.18\linewidth]{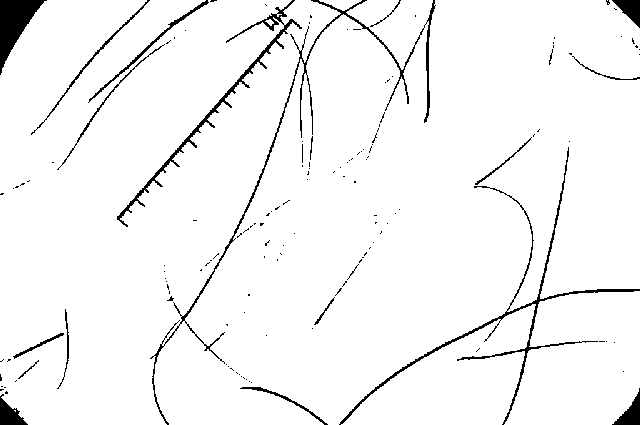} &
\includegraphics[width=0.18\linewidth]{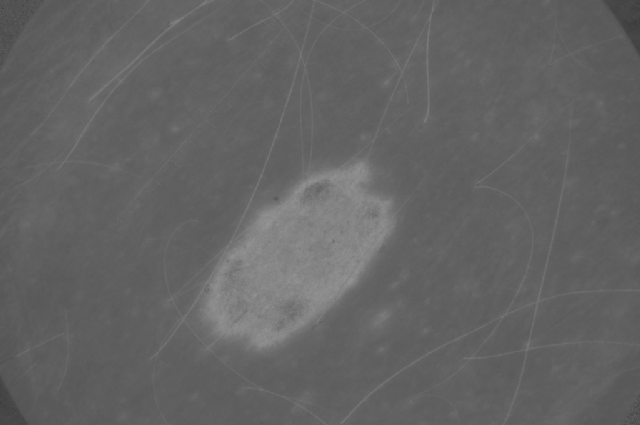} &
\includegraphics[width=0.18\linewidth]{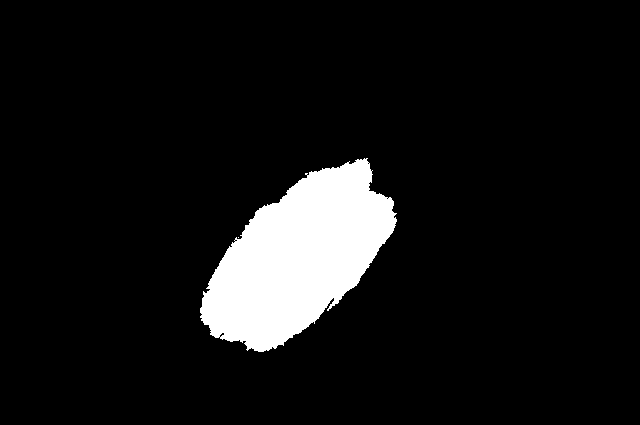} &
\includegraphics[width=0.18\linewidth]{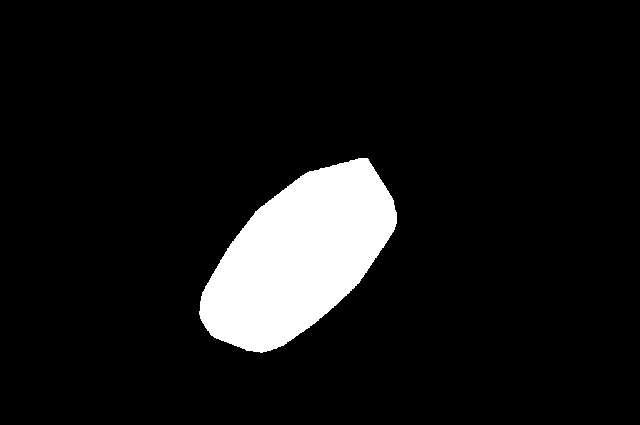}\\

\includegraphics[width=0.18\linewidth]{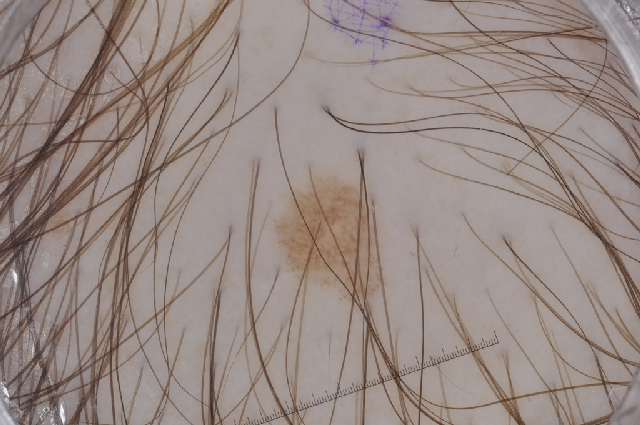} &
\includegraphics[width=0.18\linewidth]{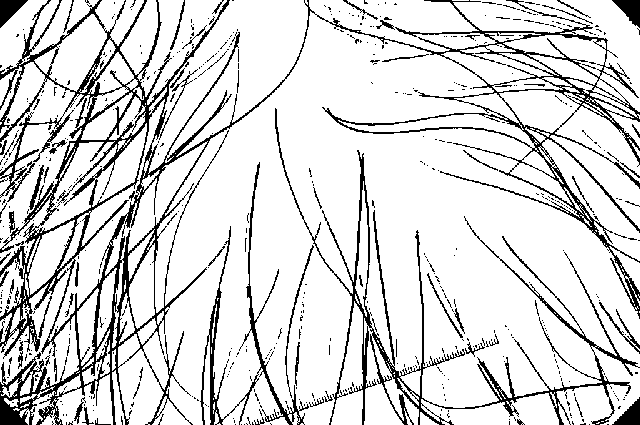} &
\includegraphics[width=0.18\linewidth]{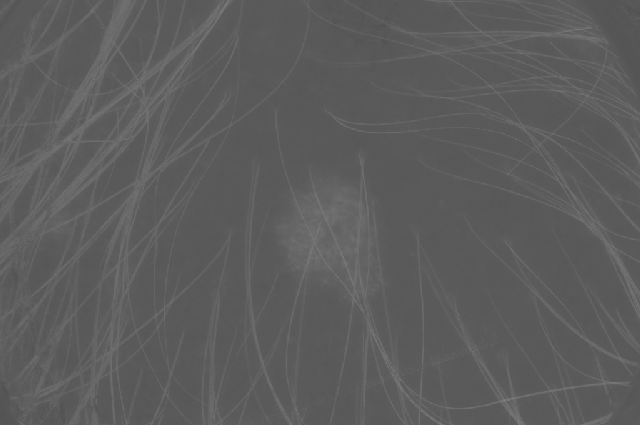} &
\includegraphics[width=0.18\linewidth]{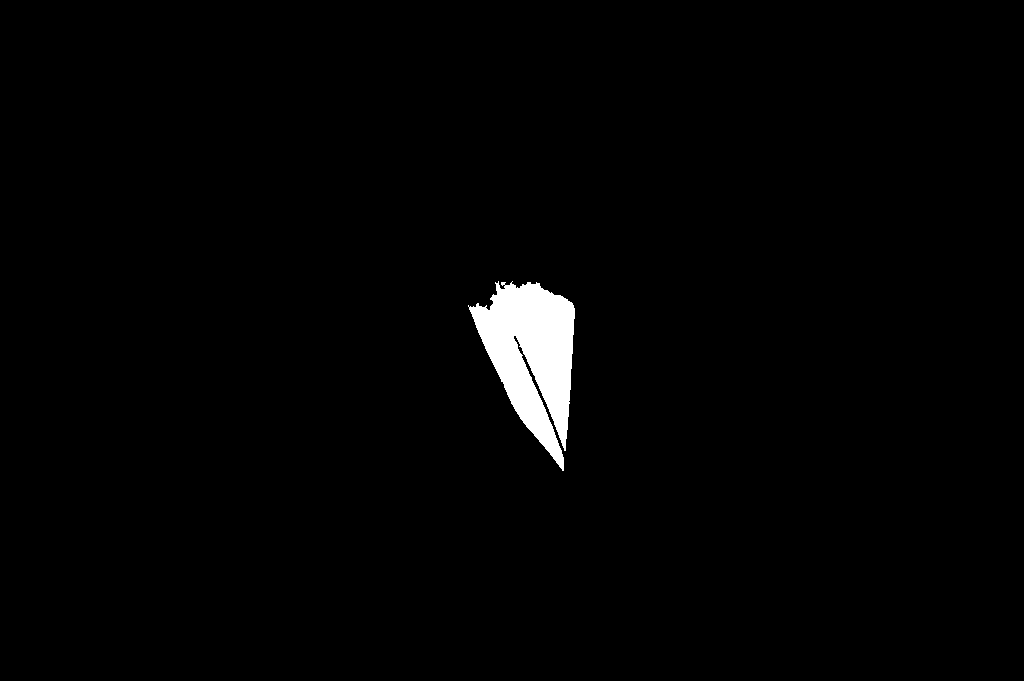} &
\includegraphics[width=0.18\linewidth]{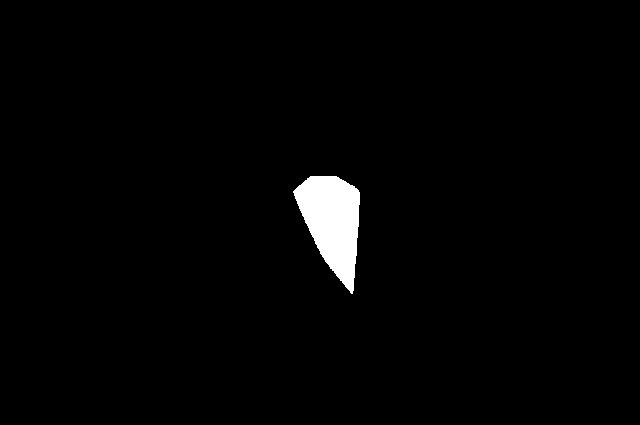}\\

(a) & (b) & (c) & (d) & (e)\\
\end{tabular}

\caption{Analysis of results for images 12092 (first row) and
13399 (second row), including hair: (a) image; (b) Rnorm band
selected for segmentation; (c) Rnorm binarization; (d) initial SDI
segmentation; (e) final SDI segmentation.} \label{Fig:Hair}
\end{figure}

In Fig. \ref{Fig:Hair}, we report two examples showing that the
bottom-hat filter adopted for excluding hair from the image ROI
performs quite well.

\begin{figure}[!htb]
\setlength{\tabcolsep}{2pt} \centering
\begin{tabular}{ccccc}
\includegraphics[width=0.18\linewidth]{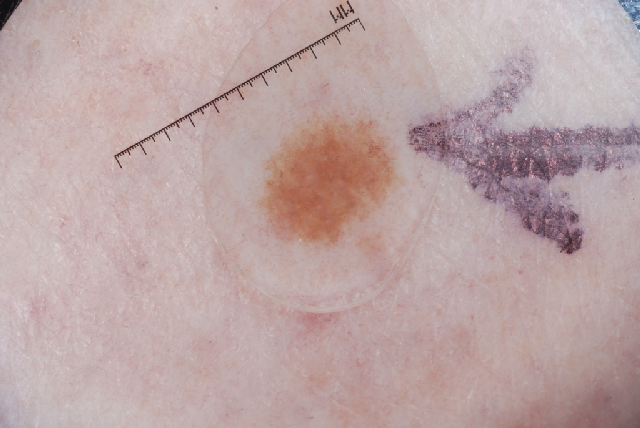} &
\includegraphics[width=0.18\linewidth]{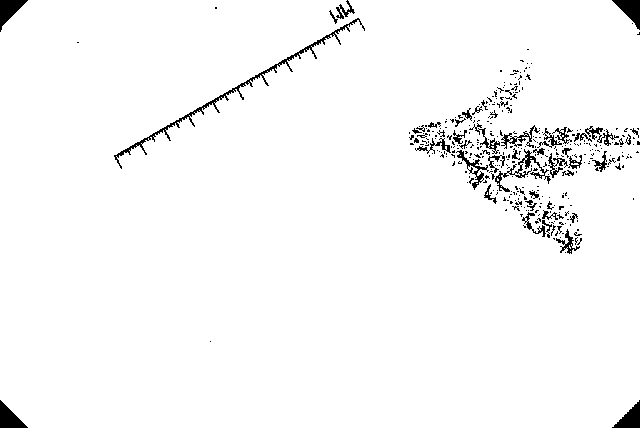} &
\includegraphics[width=0.18\linewidth]{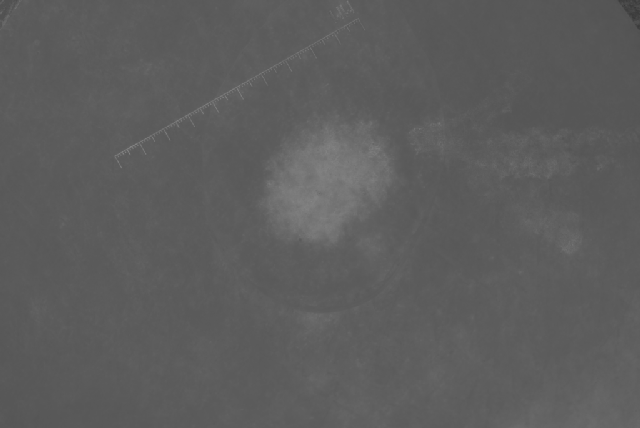} &
\includegraphics[width=0.18\linewidth]{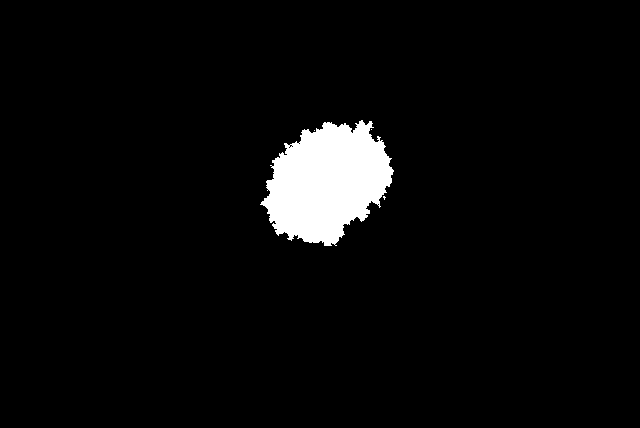} &
\includegraphics[width=0.18\linewidth]{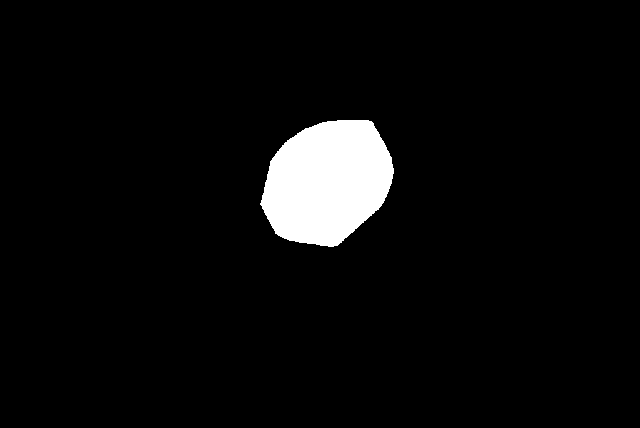}\\

\includegraphics[width=0.18\linewidth]{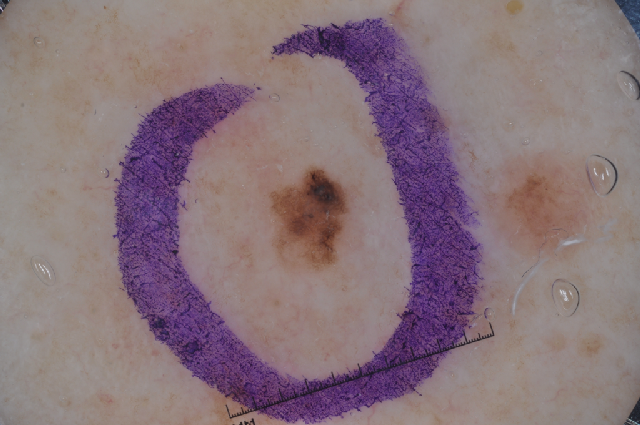} &
\includegraphics[width=0.18\linewidth]{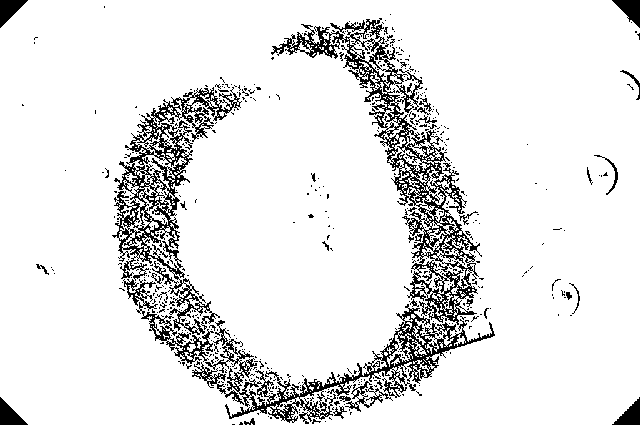} &
\includegraphics[width=0.18\linewidth]{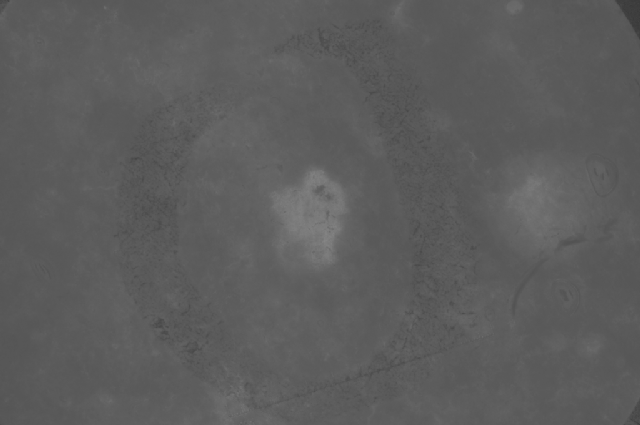} &
\includegraphics[width=0.18\linewidth]{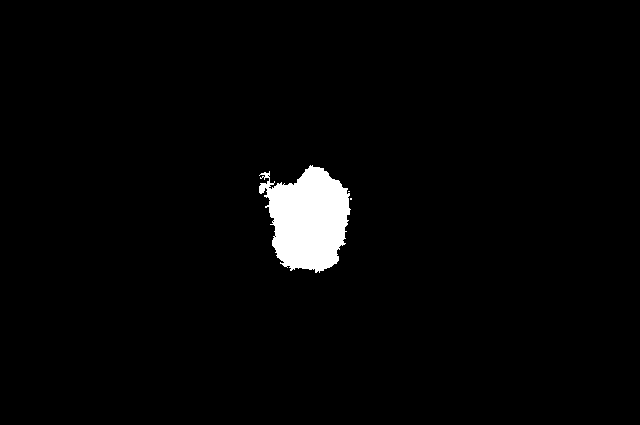} &
\includegraphics[width=0.18\linewidth]{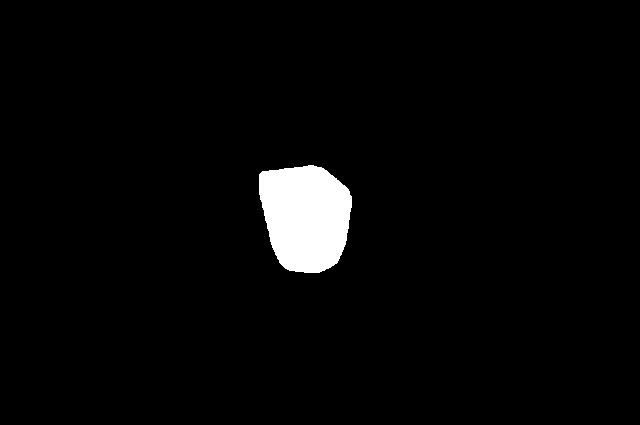}\\

(a) & (b) & (c) & (d) & (e)\\
\end{tabular}

\caption{Analysis of results for images 13216 (first row) and
13414 (second row), including ink markers: (a) image; (b) Rnorm
band selected for segmentation; (c) Rnorm binarization; (d)
initial SDI segmentation; (e) final SDI segmentation.}
\label{Fig:Ink}
\end{figure}

Fig. \ref{Fig:Ink} shows that the selection of the image ROI
partly succeeds in excluding from the ROI also ink marker signs.
Moreover, the remaining ink pixels that are included into the ROI
do not affect the final segmentation, thanks to the selection of
the Rnorm band for segmentation.

\begin{figure}[!htb]
\setlength{\tabcolsep}{2pt} \centering
\begin{tabular}{ccccc}
\includegraphics[width=0.18\linewidth]{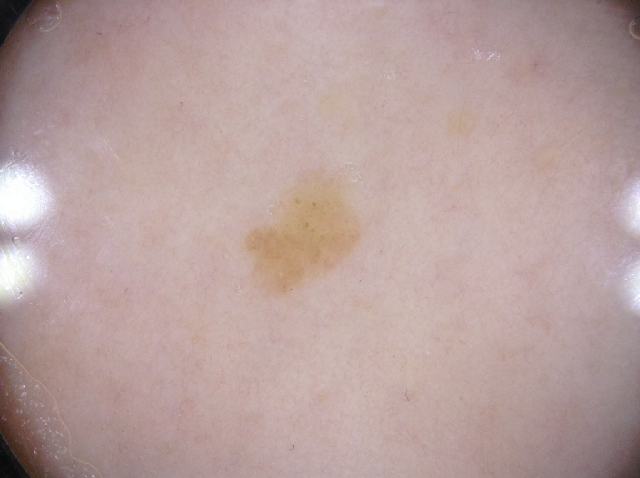} &
\includegraphics[width=0.18\linewidth]{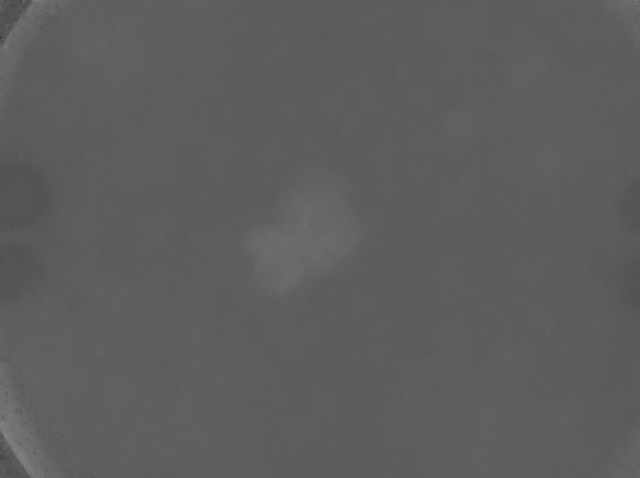}&
\includegraphics[width=0.18\linewidth]{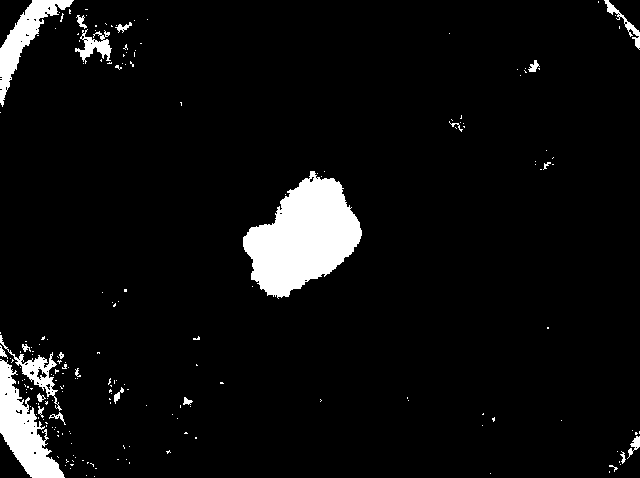} &
\includegraphics[width=0.18\linewidth]{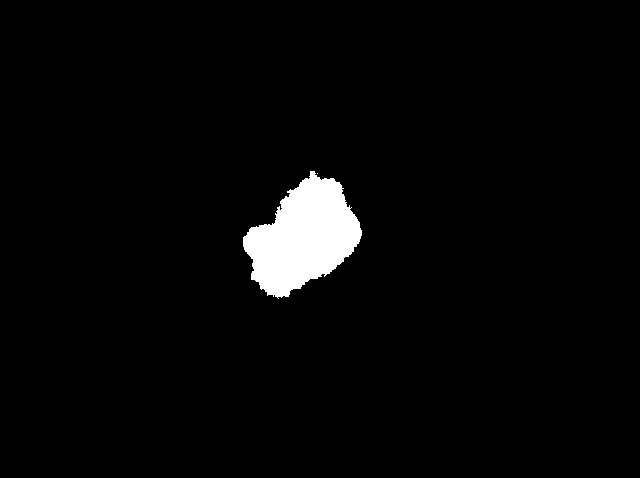} &
\includegraphics[width=0.18\linewidth]{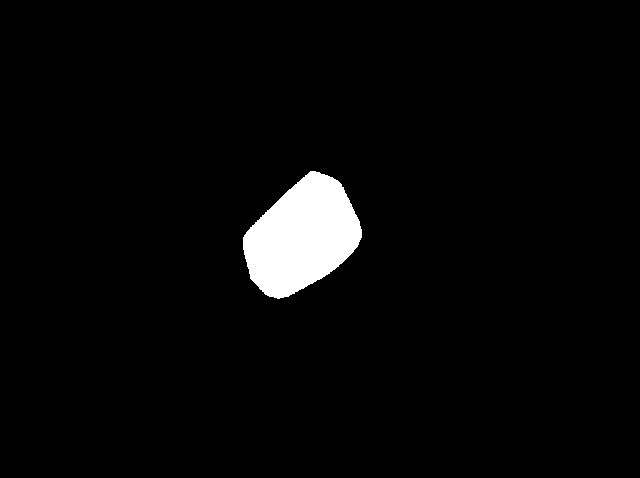}\\

\includegraphics[width=0.18\linewidth]{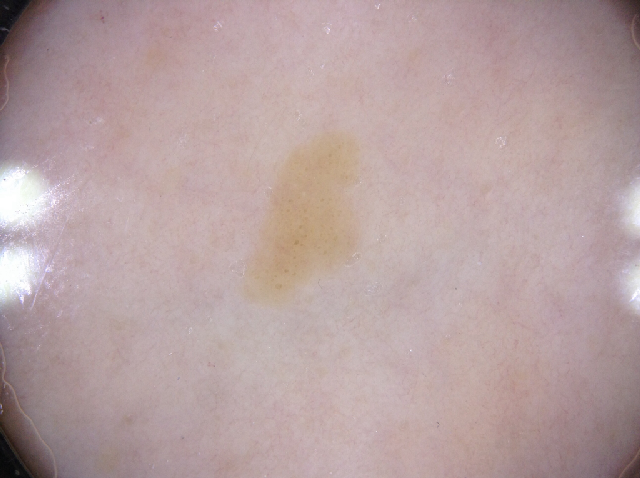} &
\includegraphics[width=0.18\linewidth]{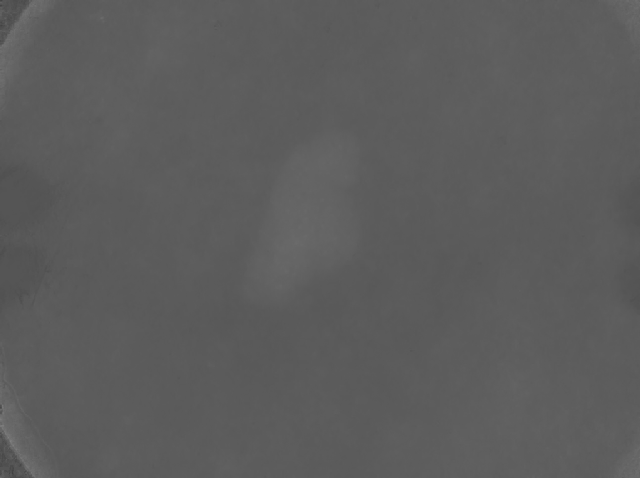} &
\includegraphics[width=0.18\linewidth]{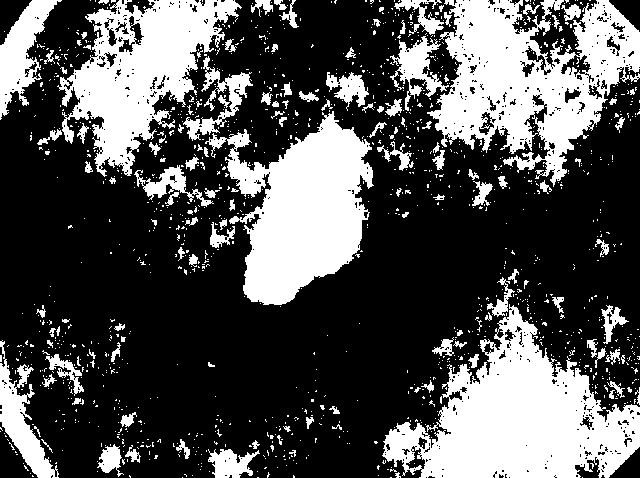} &
\includegraphics[width=0.18\linewidth]{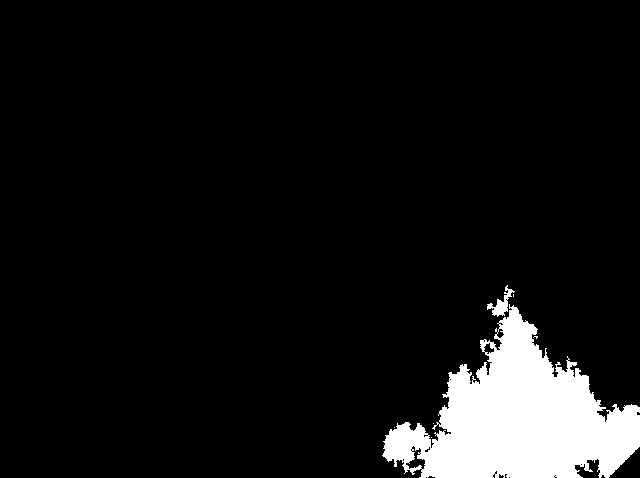} &
\includegraphics[width=0.18\linewidth]{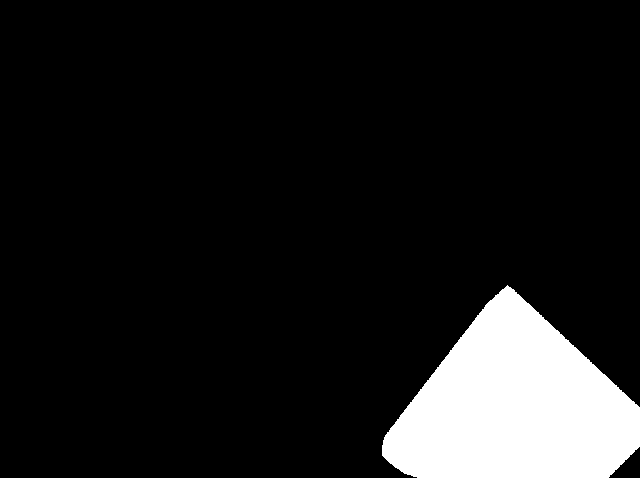}\\
(a) & (b) & (c) & (d) & (e)\\
\end{tabular}

\caption{Analysis of results for images 14574 (first row) and
14575 (second row): (a) image; (b) Rnorm band selected for
segmentation; (c) Rnorm binarization; (d) initial SDI
segmentation; (e) final SDI segmentation.} \label{Fig:Highlights}
\end{figure}

We point out the choice of the connected component having the
maximum area is not always the best for selecting a correct lesion
segmentation. An extreme example is shown in Fig.
\ref{Fig:Highlights}. Here, we can observe that image 14574 is
perfectly segmented, while image 14575, that looks pretty similar
to the previous one, is wrongly segmented. The error is due to a
very different binarization (Fig. \ref{Fig:Highlights}-(c)),
leading to the wrong connected component.

\begin{figure}[!htb]
\begin{center}
\begin{tabular}{ccc}
\includegraphics[width=0.28\linewidth]{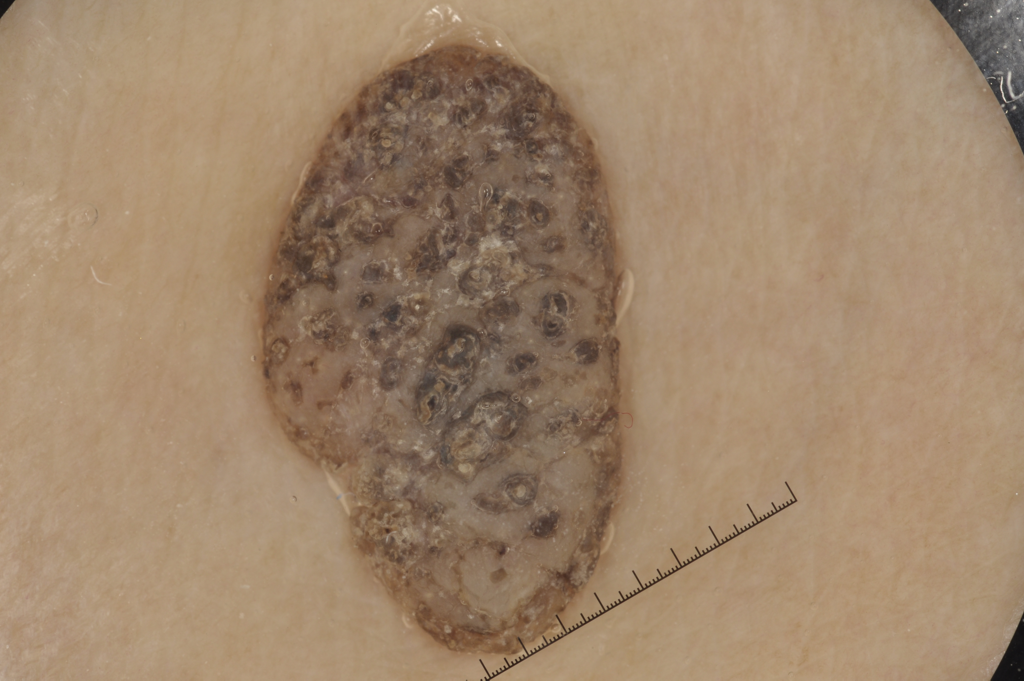} &
\includegraphics[width=0.28\linewidth]{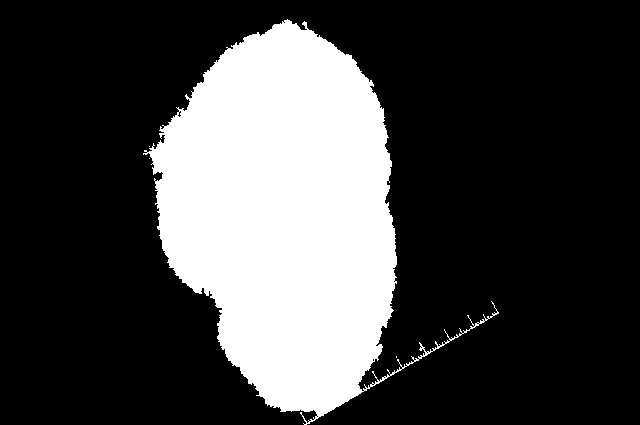} &
\includegraphics[width=0.28\linewidth]{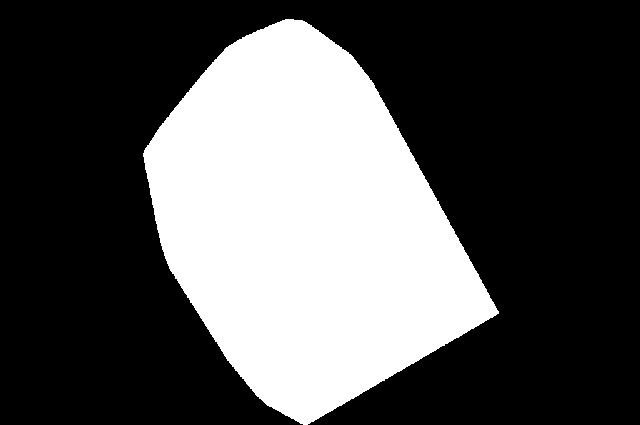}\\
(a) & (b) & (c)\\
\end{tabular}
\end{center}
\caption{Analysis of results for image 12272: (a) image; (b)
initial SDI segmentation; (c) final SDI segmentation.}
\label{Fig:CH}
\end{figure}

Finally, we observe that, although most of the times the convex
hull of the initial SDI segmentation better conforms to the ground
truth (see Fig. \ref{Fig:SegmentationCH}), it can erroneously
include into the final segmentation also wide skin areas. This is
the case, for example, of image 12272 (Fig. \ref{Fig:CH}), where
the ruler mark has been erroneously included into the initial SDI
segmentation, leading to a too wide convex hull in the final
result.

\section{Conclusions} \label{Conclusioni}

We proposed the SDI algorithm for dermoscopic image segmentation,
consisting of three main steps: selection of the image ROI,
selection of the segmentation band, and segmentation.
The reported analysis of experimental results achieved by the SDI
algorithm on the ISIC 2017 dataset allowed us to highlight its
pro's and con's.
This leads us to conclude that, although some accurate results can
be achieved, there is room for improvements in different
directions, that we will go through in future investigations.

\section*{Acknowledgments}
This research was supported by LAB GTP Project, funded by MIUR.


\end{document}